\documentclass[runningheads]{llncs}
\usepackage{graphicx}
\usepackage{comment}
\usepackage{amsmath,amssymb} 
\usepackage{color}
\usepackage{setspace}

\usepackage{booktabs}
\usepackage{setspace}
\usepackage{multirow}
\usepackage{amsmath}
\usepackage{amssymb}
\usepackage{subfig,caption}
\usepackage{array}
\usepackage{threeparttable}
\usepackage[colorlinks,linkcolor=black,anchorcolor=black,citecolor=black,urlcolor=black]{hyperref}

\newcommand{\PreserveBackslash}[1]{\let\temp=\\#1\let\\=\temp}
\newcolumntype{C}[1]{>{\PreserveBackslash\centering}p{#1}}
\makeatother

\begin{document}
	
	\pagestyle{headings}
	\mainmatter
	\def\ECCVSubNumber{692}  
	
	\title{Learning to Combine: Knowledge Aggregation for Multi-Source Domain Adaptation} 

	\titlerunning{Learning to Combine}
	\author{Hang Wang\inst{}\thanks{Equal contribution.}\and
		Minghao Xu\inst{\star}\and
		Bingbing Ni\inst{}\thanks{Corresponding author: Bingbing Ni.} \and
		Wenjun Zhang\inst{}}
	\authorrunning{Wang et al.}
	
	\institute{Shanghai Jiao Tong University, Shanghai 200240, China \\
		\email{\{wang--hang, xuminghao118, nibingbing, zhangwenjun\}@sjtu.edu.cn}}
	
	\maketitle
	
	\begin{abstract}
		Transferring knowledges learned from multiple source domains to target domain is a more practical and challenging task than conventional single-source domain adaptation. Furthermore, the increase of modalities brings more difficulty in aligning feature distributions among multiple domains. To mitigate these problems, we propose a Learning to Combine for Multi-Source Domain Adaptation (LtC-MSDA) framework via exploring interactions among domains. In the nutshell, a knowledge graph is constructed on the prototypes of various domains to realize the information propagation among semantically adjacent representations. On such basis, a graph model is learned to predict query samples under the guidance of correlated prototypes. In addition, we design a Relation Alignment Loss (RAL) to facilitate the consistency of categories' relational interdependency and the compactness of features, which boosts features' intra-class invariance and inter-class separability. Comprehensive results on public benchmark datasets demonstrate that our approach outperforms existing methods with a remarkable margin. Our code is available at \url{https://github.com/ChrisAllenMing/LtC-MSDA}.
		
	\end{abstract}
	
	
	\section{Introduction}\label{sec1}
	
	Deep Neural Network (DNN) is expert at learning discriminative representations under the support of massive labeled data, and it has achieved incredible successes in many computer-vision-related tasks, \emph{e.g.} object classification \cite{alexnet,resnet}, object detection \cite{faster_rcnn,ssd} and semantic segmentation \cite{deeplab,mask_rcnn}. However, when directly deploying the model trained on a specific dataset to the scenarios with distinct backgrounds, weather or illumination, undesirable performance decay commonly occurs, due to the existence of domain shift \cite{deep_transferrable}. 
	
	Unsupervised Domain Adaptation (UDA) is an extensively explored technique to address such problem, and it focuses on the transferability of knowledge learned from a labeled dataset (source domain) to another unlabeled one (target domain). The basic intuition behind these attempts is that knowledge transfer can be achieved by boosting domain-invariance of feature representations from different domains. In order to realize such goal, various strategies have been proposed, including minimizing explicitly defined domain discrepancy metrics \cite{dan,w-mmd,deepcoral}, adversarial-training-based domain confusion \cite{revgrad,adda,cdan} and GAN-based domain alignment \cite{pixel-level,reconstruction-classification,gta}. 
	
	However, in real-world applications, it is unreasonable to deem that the labeled images are drawn from a single domain. Actually, these samples can be collected under different deployment environments, \emph{i.e.} from multiple domains, which reflect distinct modal information. Integrating such factor into domain alignment, a more practical problem is \emph{Multi-Source Domain Adaptation} (MSDA), which dedicates to transfer the knowledges learned from multiple source domains to an unlabeled target domain. 
	
	Inspired by the theoretical analysis \cite{msda_theory_nips08,msda_theory_nips18}, recent works \cite{DCTN,M3SDA,MDDA} predict target samples by combining the predictions of source classifiers. However, the interaction of feature representations learned from different domains has not been explored to tackle MSDA tasks. Compared to combining classifiers' predictions using hand-crafted or model-induced weights, knowledge propagation among multiple domains enables related feature representations to interact with each other before final prediction, which makes the operation of domain combination learnable. In addition, although category-level domain adaptation has been extensively studied in the literature, \emph{e.g.} maximizing dual classifiers discrepancy \cite{max_discrepancy,sliced_w-distance} and prototype-based alignment \cite{semantic,transferrable_proto}, the relationships among categories are not constrained in these works. For instance, the source domain's knowledges that truck is more similar to car than person should also be applicable to target domain. Motivated by these limitations, we propose a novel framework and loss function for MSDA as follows.
	
	
	\begin{figure}[t]
		\centering
		\setlength{\belowcaptionskip}{-0.4cm}
		\includegraphics[width=0.98\textwidth]{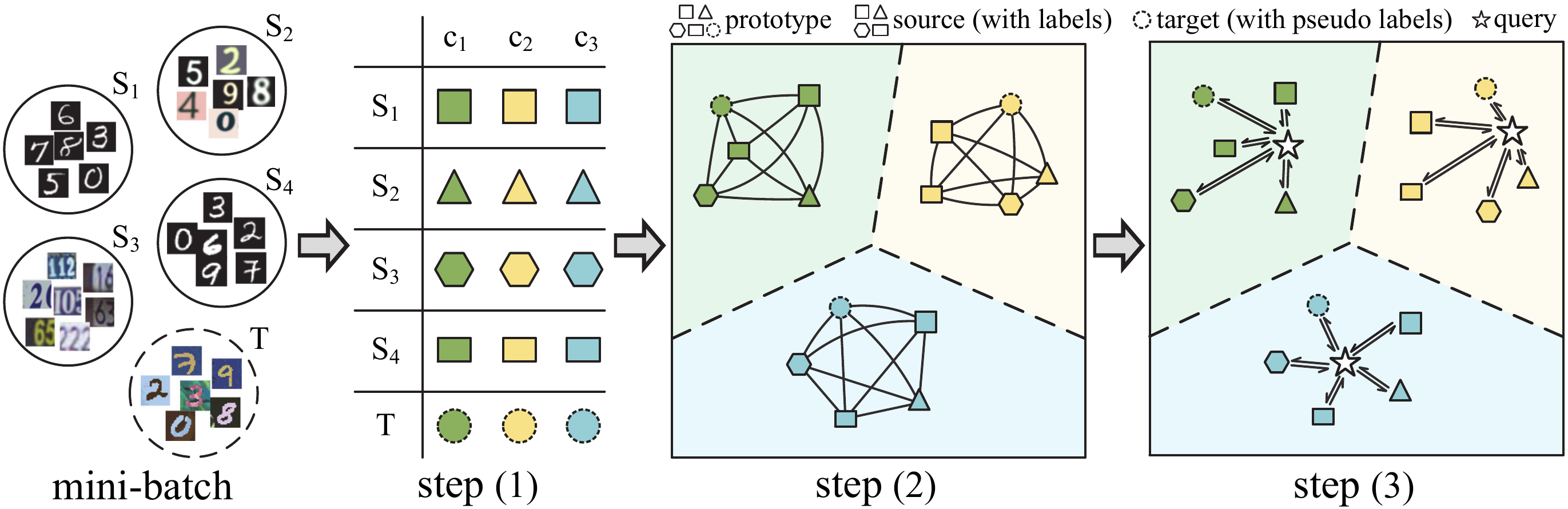}
		\caption{Given a randomly sampled mini-batch, in step (1), our model first updates each category's global prototype for all domains. In step (2), a knowledge graph is constructed on these prototypes. Finally, in step (3), a bunch of query samples are inserted into the graph and predicted via knowledge aggregation. } 
		\label{fig_motivation}
	\end{figure}
	
	
	\textbf{Learning to Combine.} We propose a new framework, \emph{Learning to Combine for MSDA} (LtC-MSDA), which leverages the knowledges learned from multiple source domains to assist model's inference on target domain. In the training phase, three major steps are performed, which are graphically illustrated in Figure~\ref{fig_motivation}. (1) \emph{Global prototype\footnotemark[5]\footnotetext[5]{Prototype is the mean embedding of all samples within the same class.} maintenance}: Based on a randomly sampled mini-batch containing samples from source and target domains, we estimate the prototype representation of each category for all domains. In order to mitigate the randomness of these estimations, global prototypes are maintained through a moving average scheme. (2) \emph{Knowledge graph construction}: In this step, a knowledge graph is constructed on the global prototypes of different domains, and the connection weight between two global prototypes is determined by their similarity. (3) \emph{Knowledge-aggregation-based prediction}: Given a bunch of query samples from arbitrary domains, we first extend the knowledge graph with these samples. After that, a graph convolutional network (GCN) is employed to propagate feature representations throughout the extended graph and output the classification probability for each node. After training, the knowledge graph is saved, and only step (3) is conducted for model's inference. 
	
	\textbf{Class-relation-aware Domain Alignment.} During the process of domain adaptation, in order to exploit the relational interdependency among categories, we propose a \emph{Relation Alignment Loss} (RAL), which is composed of a global and a local term. (1) \emph{Global relation constraint}: In this term, based on the adjacency matrix of knowledge graph, we constrain the connection weight between two arbitrary classes to be consistent among all domains, which refines the relative position of different classes' features in the latent space. (2) \emph{Local relation constraint}: This term facilitates the compactness of various categories' features. In specific, we restrain the feature representation of a sample to be as close as possible to its corresponding global prototype, which makes the features belonging to distinct categories easier to be separated.
	
	Our contributions can be summarized as follows:
	\begin{enumerate}
		\item We propose a Learning to Combine for MSDA (LtC-MSDA) framework, in which the knowledges learned from source domains interact with each other and assist model's prediction on target domain. 
		\item In order to better align the feature distributions of source and target domains,
		we design a Relation Alignment Loss (RAL) to constrain the global and local relations of feature representations. 
		\item We evaluate our model on three benchmark datasets with different domain shift and data complexity, and extensive results show that the proposed method outperforms existing approaches with a clear margin. 
	\end{enumerate}
	
	
	\section{Related Work} \label{sec2}
	
	\par{\textbf{Unsupervised Domain Adaptation (UDA).}}
	UDA seeks to generalize a model learned from a labeled source domain to a new target domain without labels. Many previous methods achieve such goal via minimizing an explicit domain discrepancy metric \cite{domain_confusion,w-mmd,dan,sliced_w-distance,deepcoral}. Adversarial learning is also employed to align two domains on feature level \cite{revgrad,adda,cdan} or pixel level \cite{pixel-level,reconstruction-classification,gta,domain_mixup}.
	Recently, a group of approaches performs category-level domain adaptation through utilizing dual classifier \cite{max_discrepancy,sliced_w-distance}, or domain prototype \cite{semantic,transferrable_proto,cross_domain_detection}.
	In this work, we further explore the consistency of category relations on all domains. 
	
	
	\par{\textbf{Multi-Source Domain Adaptation (MSDA).}}
	MSDA assumes data are collected from multiple source domains with different distributions, which is a more practical scenario compared to single-source domain adaptation. 
	Early theoretical analysis \cite{msda_theory_nips08,bounds_DA} gave strong guarantees for representing target distribution as the weighted combination of source distributions. 
	Based on these works, Hoffman \emph{et al.} \cite{msda_theory_nips18} derived normalized solutions for MSDA problems. 
	Recently, Zhao \emph{et al.} \cite{MDAN} aligned target domain to source domains globally using adversarial learning. Xu \emph{et al.} \cite{DCTN} deployed multi-way adversarial learning and combined source-specific perplexity scores for target predictions. Peng \emph{et al.} \cite{M3SDA} proposed to transfer knowledges by matching the moments of feature representations. In \cite{MDDA}, source distilling mechanism is introduced to fine-tune the separately pre-trained feature extractor and classifier.
	
	\emph{Improvements over existing methods.} In order to derive the predictions of target samples, former works \cite{DCTN,M3SDA,MDDA} utilize the ensemble of source classifiers to output weighted classification probabilities, while such combination scheme prohibits the end-to-end learnable model. In this work, we design a \emph{Learning to Combine} framework to predict query samples based on the interaction of knowledges learned from source and target domains, which makes the whole model end-to-end learnable. 
	
	
	\par{\textbf{Knowledge Graph.}}
	A knowledge graph describes entities and their interrelations, organized in a graph. 
	Learning knowledge graphs and using attribute relationships has recently been of interest to the vision community. 
	Several works \cite{kg_nlu_Hakkani,kg_nlp_Krishnamurthy} utilize knowledge graphs based on the defined semantic space for natural language understanding.
	For multi-label image classification \cite{kg_cls_Marino,kg_zsl_lee}, knowledge graphs are applied to exploit explicit semantic relations. 
	In this paper, we construct a knowledge graph on global prototypes of different domains, which lays foundation for our method. 
	
	
	\par{\textbf{Graph Convolutional Network (GCN).}}
	GCN \cite{gcn_model} is designed to compute directly on graph-structured data and model the inner structural relations. 
	Such structures typically come from some prior knowledges about specific problems. 
	Due to its effectiveness, GCNs have been widely used in various tasks, \emph{e.g.} action recognition \cite{graph_action_recognition}, person Re-ID \cite{graph_reid_yan,graph_reid_liu} and point cloud learning \cite{graph_point_cloud}. For MSDA task, we employ GCN to propagate information on the knowledge graph.
	
	
	\section{Method} \label{sec3}
	
	In Multi-Source Domain Adaptation (MSDA), there are $M$ source domains $S_1$, $S_2$, $\cdots$, $S_M$. The domain $S_m = \{(x^{\mathcal{S}_m}_i, y^{\mathcal{S}_m}_i)\}^{N_{\mathcal{S}_m}}_{i=1}$ is characterized by $N_{\mathcal{S}_m}$ i.i.d. labeled samples, where $x^{\mathcal{S}_m}_i$ follows one of the source distributions $\mathbb{P}_{\mathcal{S}_m}$ and $y^{\mathcal{S}_m}_i \in \{1, 2, \cdots, K\}$ ($K$ is the number of classes) denotes its corresponding label. Similarly, target domain $\mathcal{T} = \{x^\mathcal{T}_j\}^{N_\mathcal{T}}_{j=1}$ is represented by $N_\mathcal{T}$ i.i.d. unlabeled samples, where $x^\mathcal{T}_j$ follows target distribution $\mathbb{P}_\mathcal{T}$. In the training phase, a randomly sampled mini-batch $B = \{ \widehat{\mathcal{S}}_1, \widehat{\mathcal{S}}_2, \cdots , \widehat{\mathcal{S}}_M, \widehat{\mathcal{T}} \}$ is used to characterize 
	source and target domains, and $|B|$ denotes the batch size. 
	
	\subsection{Motivation and Overview} \label{sec3_1}
	
	For MSDA, the core research topic is how to achieve more precise predictions for target samples through fully utilizing the knowledges among different domains. 
	In order to mitigate the error of single-source prediction, recent works \cite{DCTN,M3SDA,MDDA} express the classification probabilities of target samples as the weighted average of source classifiers' predictions. However, such scheme requires prior knowledges about the relevance of different domains to obtain combination weights, which makes the whole model unable to be end-to-end learnable. 
	
	In addition, learning to generalize from multiple source domains to target domain has a ``double-edged sword'' effect on model's performance.
	From one perspective, samples from multiple domains provide more abundant modal information of different classes, and thus the decision boundaries are refined according to more support points. From the other perspective, the distribution discrepancy among distinct source domains increases the difficulty of learning domain-invariant features. Off-the-shelf UDA techniques might fail in the condition that multi-modal distributions are to be aligned, since the relevance among different modalities, \emph{i.e.} categories of various domains, are not explicitly constrained in these methods. Such constraints \cite{contrastive_loss,triplet_loss} are proved to be necessary when large amounts of clusters are formed in the latent space. 
	
	To address above issues, we propose a \emph{Learning to Combine for MSDA} (LtC-MSDA) framework. In specific, a knowledge graph is constructed on the prototypes of different domains to enable the interaction among semantically adjacent entities, and query samples are added into this graph to obtain their classification probabilities under the guidance of correlated prototypes. In this process, the combination of different domains' knowledges is achieved via information propagation, which can be learned by a graph model. On the basis of this framework, a \emph{Relation Alignment Loss} (RAL) is proposed, which facilitates the consistency of categories' relational interdependency on all domains and boosts the compactness of feature embeddings within the same class.
	
	
	\begin{figure}[t]
		\centering
		\setlength{\belowcaptionskip}{-0.4cm}
		\includegraphics[width=0.98\textwidth]{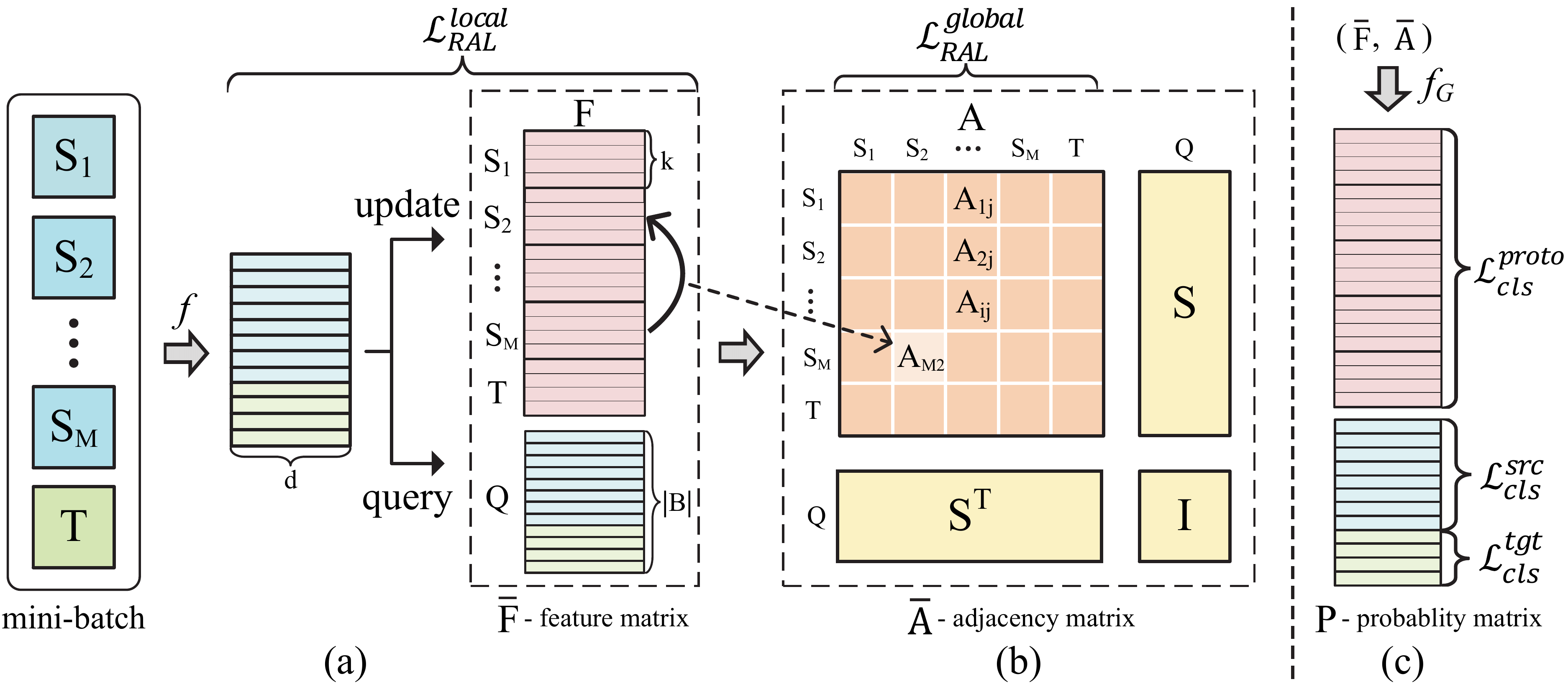}
		\caption{\textbf{Framework overview.} (a) A randomly sampled mini-batch is utilized to update global prototypes and also serves as query samples, and the local relation loss $\mathcal{L}^{local}_{RAL}$ is constrained to promote feature compactness. (b) A knowledge graph is constructed on prototypes, whose adjacency matrix $\textbf{A}$ embodies the relevance among different domains' categories. On the basis of block matrices in $\textbf{A}$, global relation loss $\mathcal{L}^{global}_{RAL}$ is derived. (c) Extended by query samples, feature matrix $\bar{\textbf{F}}$ and adjacency matrix $\bar{\textbf{A}}$ are fed into a GCN model $f_G$ to produce final predictions $\textbf{P}$. On such basis, three kinds of classification losses are defined. } 
		\label{fig_framework}
	\end{figure}
	
	
	\subsection{Learning to Combine for MSDA} \label{sec3_2}
	
	In the proposed LtC-MSDA framework, for each training iteration, a mini-batch containing samples from all domains is mapped to latent space, and the produced feature embeddings are utilized to update global prototypes and also served as queries. After that, global prototypes and query samples are structured as a knowledge graph. Finally, a GCN model is employed to perform information propagation and output classification probability for each node of knowledge graph. 
	Figure~\ref{fig_framework} gives a graphical illustration of the whole framework, and its details are presented in the following parts.
	
	\textbf{Global prototype maintenance.} This step updates global prototypes with mini-batch statistics. Based on a mini-batch $B$, we estimate the prototype of each category for all domains. For source domain $\mathcal{S}_m$, the estimated prototype $\widehat{c}^{\mathcal{S}_m}_{k}$ is defined as the mean embedding of all samples belonging to class $k$ in $\widehat{\mathcal{S}}_m$:
	\begin{equation} \label{eq1}
		\widehat{c}^{\mathcal{S}_m}_{k} = \frac{1}{| \widehat{\mathcal{S}}^k_{m} |} \sum_{(x^{\mathcal{S}_m}_i, y^{\mathcal{S}_m}_i) \in \widehat{\mathcal{S}}^k_{m}} f(x^{\mathcal{S}_m}_i) ,
	\end{equation}
	where $\widehat{\mathcal{S}}^k_m$ is the set of all samples with class label $k$ in the sampling $\widehat{\mathcal{S}}_m$, and $f$ represents the mapping from image to feature embedding.
	
	For target domain $\mathcal{T}$, since ground truth information is unavailable, we first assign pseudo labels for the samples in $\widehat{\mathcal{T}}$ using the strategy proposed by \cite{collaborative}, and the estimated prototype $\widehat{c}^{\mathcal{T}}_{k}$ of target domain is defined as follows:
	\begin{equation} \label{eq2}
		\widehat{c}^{\mathcal{T}}_{k} = \frac{1}{| \widehat{\mathcal{T}}_k |} \sum_{(x^{\mathcal{T}}_i, \widehat{y}^{\mathcal{T}}_i) \in \widehat{\mathcal{T}}_k} f(x^{\mathcal{T}}_i) ,
	\end{equation}
	where $\widehat{y}^{\mathcal{T}}_i$ is the pseudo label assigned to $x^{\mathcal{T}}_i$, and $\widehat{\mathcal{T}}_k$ denotes the set of all samples labeled as class $k$ in $\widehat{\mathcal{T}}$. 
	In order to correct estimation bias brought by the randomness of mini-batch samplings, we maintain the global prototypes for source and target domains with an exponential moving average scheme: 
	\begin{equation} \label{eq3}
		c^{\mathcal{S}_m}_{k} := \beta c^{\mathcal{S}_m}_{k} + (1 - \beta) \widehat{c}^{\mathcal{S}_m}_{k} \quad \ m = 1, 2, \cdots , M , 
	\end{equation}
	\begin{equation} \label{eq4}
		c^{\mathcal{T}}_{k} := \beta c^{\mathcal{T}}_{k} + (1 - \beta) \widehat{c}^{\mathcal{T}}_{k} ,
	\end{equation}
	where $\beta$ is the exponential decay rate which is fixed as $0.7$ in all experiments.
	Such moving average scheme is broadly used in the literature \cite{adam,semantic,moco} to stabilize the training process through smoothing global variables. 
	
	
	\textbf{Knowledge graph construction.} In order to further refine category-level representations with knowledges learned from multiple domains, this step structures the global prototypes of various domains as a knowledge graph $\mathcal{G} = (\mathcal{V}, \mathcal{E})$. In this graph, the vertex set $\mathcal{V}$ corresponds to $(M+1)K$ prototypes, and the feature matrix $\textbf{F} \in \mathbb{R}^{| \mathcal{V} | \times d}$ ($d$: the dimension of feature embedding) is defined as the concatenation of global prototypes: 
	\begin{equation} \label{eq5}
		\textbf{F} = \Big [ \underbrace{c^{\mathcal{S}_1}_1 \, c^{\mathcal{S}_1}_2 \, \cdots \, c^{\mathcal{S}_1}_K}_{\textrm{prototypes of } \mathcal{S}_1} \ \cdots \ \underbrace{c^{\mathcal{S}_M}_1 \, c^{\mathcal{S}_M}_2 \, \cdots \, c^{\mathcal{S}_M}_K}_{\textrm{prototypes of } \mathcal{S}_M} \ \underbrace{c^{\mathcal{T}}_1 \, c^{\mathcal{T}}_2 \, \cdots \, c^{\mathcal{T}}_K}_{\textrm{prototypes of } \mathcal{T}} \Big ]^{\mathrm{T}} .
	\end{equation}
	
	The edge set $\mathcal{E} \subseteq \mathcal{V} \times \mathcal{V}$ describes the relations among vertices, and an adjacency matrix $\textbf{A} \in \mathbb{R}^{| \mathcal{V} | \times | \mathcal{V} |}$ is employed to model such relationships. 
	In specific, we derive the adjacency matrix by applying a Gaussian kernel $\mathcal{K}_G$ over pairs of global prototypes:
	\begin{equation} \label{eq6}
		\textbf{A}_{i,j} = \mathcal{K}_G (\textbf{F}^{\mathrm{T}}_i, \textbf{F}^{\mathrm{T}}_j) = \textrm{exp} \Big ( - \frac{|| \textbf{F}^{\mathrm{T}}_i - \textbf{F}^{\mathrm{T}}_j ||^2_2}{2 \sigma^2} \Big ) ,  
	\end{equation}
	where $\textbf{F}^{\mathrm{T}}_i$ and $\textbf{F}^{\mathrm{T}}_j$ denote the $i$-th and $j$-th global prototype in feature matrix $\textbf{F}$, and $\sigma$ is the standard deviation parameter controlling the sparsity of $\textbf{A}$.
	
	\textbf{Knowledge-aggregation-based prediction.} In this step, we aim to obtain more accurate predictions for query samples under the guidance of multiple domains' knowledges. We regard the mini-batch $B$ as a bunch of query samples and utilize them to establish an extended knowledge graph $\bar{\mathcal{G}} = (\bar{\mathcal{V}}, \bar{\mathcal{E}})$. In this graph, the vertex set $\bar{\mathcal{V}}$ is composed of the original vertices in $\mathcal{V}$, \emph{i.e.} global prototypes, and query samples' feature embeddings, which yields an extended feature matrix $\bar{\textbf{F}} \in \mathbb{R}^{| \bar{\mathcal{V}} | \times d}$ as follows: 
	\begin{equation} \label{eq7}
		\bar{\textbf{F}} = \Big [ \textbf{F}^{\mathrm{T}} \; f(q_1) \; f(q_2) \; \cdots \; f(q_{| B |}) \Big ]^{\mathrm{T}} , 
	\end{equation}
	where $q_i$ ($i=1,2, \cdots, | B |$) denotes the $i$-th query sample. 
	
	The edge set $\bar{\mathcal{E}}$ is expanded with the edges of new vertices. Concretely, an extended adjacency matrix $\bar{\textbf{A}}$ is derived by adding the connections between global prototypes and query samples:
	\begin{equation} \label{eq8}
		\textbf{S}_{i,j} = \mathcal{K}_G (\textbf{F}^{\mathrm{T}}_i, f(q_j)) = \textrm{exp} \Big ( - \frac{|| \textbf{F}^{\mathrm{T}}_i - f(q_j) ||^2_2}{2 \sigma^2} \Big ) ,
	\end{equation}
	\begin{equation} \label{eq9}
		\bar{\textbf{A}} = \left[
		\begin{array}{cc}
			\textbf{A} \ & \textbf{S} 
			\\ 
			\textbf{S}^\mathrm{T} \ & \textbf{I}
		\end{array}
		\right] ,
	\end{equation}
	where $\textbf{S} \in \mathbb{R}^{| \mathcal{V} | \times | B |}$ is the similarity matrix measuring the relevance between original and new vertices. Considering that the semantic information from a single sample is not precise enough, we ignore the interaction among query samples and use an identity matrix $\textbf{I}$ to depict their relations. 
	
	After these preparations, a Graph Convolutional Network (GCN) is employed to propagate feature representations throughout the extended knowledge graph, such that the representations within the same category are encouraged to be consistent across all domains and query samples. In specific, inputted with the feature matrix $\bar{\textbf{F}}$ and adjacency matrix $\bar{\textbf{A}}$, the GCN model $f_G$ outputs the classification probability matrix $\textbf{P} \in \mathbb{R}^{| \bar{\mathcal{V}} | \times K}$ as follows:
	\begin{equation} \label{eq10}
		\textbf{P} = f_G \, (\bar{\textbf{F}}, \; \bar{\textbf{A}}) .
	\end{equation}
	
	\emph{Model inference.} After training, we store the feature extractor $f$, GCN model $f_G$, feature matrix $\textbf{F}$ and adjacency matrix $\textbf{A}$. For inference, only the \emph{knowledge-aggregation-based prediction} step  is conducted. Concretely, based on the feature embeddings extracted by $f$, the extended feature matrix $\bar{\textbf{F}}$ and adjacency matrix $\bar{\textbf{A}}$ are derived by Eq.~\ref{eq7} and Eq.~\ref{eq9} respectively. Using these two matrices, the GCN model $f_G$ produces the classification probabilities for test samples. 
	
	
	\subsection{Class-relation-aware Domain Alignment}
	\label{sec3_3}
	
	In the training phase, our model is optimized by two kinds of losses which facilitate the domain-invariance and distinguishability of feature representations. The details are stated below.
	
	\textbf{Relation Alignment Loss (RAL).} This loss aims to conduct domain alignment on category level. During the domain adaptation process, except for promoting the invariance of same categories' features, it is necessary to constrain the relative position of different categories' feature embeddings in the latent space, 
	especially when numerous modalities exist in the task, \emph{e.g.} MSDA. Based on this idea, we propose the RAL which consists of a global and a local constraint:
	\begin{equation} \label{eq11}
		\mathcal{L}_{RAL} = \lambda_1 \mathcal{L}^{global}_{RAL} + \lambda_2 \mathcal{L}^{local}_{RAL} ,
	\end{equation}
	where $\lambda_1$ and $\lambda_2$ are trade-off parameters.
	
	For the global term, we facilitate the relevance between two arbitrary classes to be consistent on all domains, which is implemented through measuring the similarity of block matrices in $\textbf{A}$:  
	\begin{equation} \label{eq12}
		\mathcal{L}^{global}_{RAL} = \frac{1}{(M+1)^4} \sum_{i,j,m,n=1}^{M+1} || \textbf{A}_{i,j} - \textbf{A}_{m,n} ||_{F} ,
	\end{equation}
	where the block matrix $\textbf{A}_{i,j}$ ($1 \leqslant i,j \leqslant M+1$) evaluates all categories' relevance between the $i$-th and $j$-th domain, which is shown in Figure~\ref{fig_framework}(b), and $|| \cdot ||_{F}$ denotes Frobenius norm. In this loss, features' intra-class invariance is boosted by the constraints on block matrices' main diagonal elements, and the consistency of different classes' relational interdependency is promoted by the constraints on other elements of block matrices. 
	
	For the local term, we enhance the feature compactness of each category via impelling the feature embeddings of samples in mini-batch $B$ to approach their corresponding global prototypes, which derives the following loss function:
	\begin{equation} \label{eq13}
		\begin{split}
			\mathcal{L}^{local}_{RAL} = \frac{1}{| B |} \, \sum_{k=1}^{K} \Bigg ( \sum_{m=1}^{M} & \, \sum_{(x^{\mathcal{S}_m}_i, y^{\mathcal{S}_m}_i) \in \widehat{\mathcal{S}}^k_{m}} || f(x^{\mathcal{S}_m}_i) - c^{\mathcal{S}_m}_{k} ||^2_2 \\
			& + \sum_{(x^{\mathcal{T}}_i, \widehat{y}^{\mathcal{T}}_i) \in \widehat{\mathcal{T}}_k} || f(x^{\mathcal{T}}_i) - c^{\mathcal{T}}_{k} ||^2_2 \Bigg ) .
		\end{split}
	\end{equation}
	
	
	\textbf{Classification losses.} This group of losses aims to enhance features' distinguishability. Based on the predictions of all vertices in extended knowledge graph $\bar{\mathcal{G}}$, the classification loss is defined as the composition of three terms for global prototypes, source samples and target samples respectively:
	\begin{equation} \label{eq14}
		\mathcal{L}_{cls} = \mathcal{L}^{proto}_{cls} + \mathcal{L}^{src}_{cls} + \mathcal{L}^{tgt}_{cls} .
	\end{equation}
	
	For the global prototypes and source samples, since their labels are available, two cross-entropy losses are employed for evaluation:
	\begin{equation} \label{eq15}
		\mathcal{L}^{proto}_{cls} = \frac{1}{(M+1)K} \bigg ( \sum_{m=1}^{M} \sum_{k=1}^{K} \, \mathcal{L}_{ce} \big ( p(c^{\mathcal{S}_m}_{k}), k \big ) + \sum_{k=1}^{K} \, \mathcal{L}_{ce} \big ( p(c^{\mathcal{T}}_{k}), k \big ) \bigg ) ,
	\end{equation}
	\begin{equation} \label{eq16}
		\mathcal{L}^{src}_{cls} = \frac{1}{M} \sum_{m=1}^M \, \Big ( \mathbb{E}_{(x^{\mathcal{S}_m}_i, y^{\mathcal{S}_m}_i) \in \widehat{\mathcal{S}}_{m}} \mathcal{L}_{ce} \big( p(x^{\mathcal{S}_m}_i), y^{\mathcal{S}_m}_i \big ) \Big ) ,
	\end{equation}
	where $\mathcal{L}_{ce}$ denotes the cross-entropy loss function, and $p(x)$ represents the classification probability of $x$.
	
	For the target samples, it is desirable to make their predictions more deterministic, and thus an entropy loss is utilized for measurement:
	\begin{equation} \label{eq17}
		\mathcal{L}^{tgt}_{cls} = - \mathbb{E}_{(x^{\mathcal{T}}_i, \widehat{y}^{\mathcal{T}}_i) \in \widehat{\mathcal{T}}} \sum_{k=1}^K  \, p(\widehat{y}^{\mathcal{T}}_i=k|x^{\mathcal{T}}_i) \, \mathrm{log} \, p(\widehat{y}^{\mathcal{T}}_i=k|x^{\mathcal{T}}_i) ,
	\end{equation}
	where $p(y=k|x)$ is the probability that $x$ belongs to class $k$.
	
	
	\textbf{Overall objectives.} Combining the classification and domain adaptation losses defined above, the overall objectives for feature extractor $f$ and GCN model $f_G$ are as follows:
	\begin{equation} \label{eq18}
		\min \limits_{f} \, \mathcal{L}_{cls} + \mathcal{L}_{RAL} , \qquad \min \limits_{f_G} \, \mathcal{L}_{cls} .
	\end{equation}
	
	
	\section{Experiments} \label{sec4}
	
	In this section, we first describe the experimental settings and then compare our model with existing methods on three Multi-Source Domain Adaptation datasets to demonstrate its effectiveness. 
	
	
	\subsection{Experimental Setup} \label{sec4_1}
	
	\textbf{Training details.} 
	For all experiments, a GCN model with two graph convolutional layers is employed, in which the dimension of feature representation is $d \rightarrow d \rightarrow K$ ($d$: dimension of feature embedding; $K$: number of classes). 
	The trade-off parameters $\lambda_{1}$, $\lambda_{2}$ are set as 20, 0.001 respectively, and the standard deviation $\sigma$ is set as 0.005. 
	In addition, ``$\rightarrow D$" denotes the task of transferring from other domains to domain $D$.
	Due to space limitations, more implementation details and the results on PACS\cite{pacs} dataset are provided \textit{Appendix}. 
	
	
	\begin{table}[t]
		\begin{spacing}{1.02}
			\centering
			\scriptsize
			\caption{Classification accuracy (mean $\pm$ std \%) on \emph{Digits-five} dataset.} \label{table_digit}
			\setlength{\tabcolsep}{1.6mm}
			\begin{tabular}{c|c|c|c|c|c|c|C{0.8cm}}
				\toprule[1.0pt]
				\multirow{1}{*}{Standards} & Methods & $\rightarrow$ \textbf{mm} & $\rightarrow$ \textbf{mt} &$\rightarrow$ \textbf{up} & $\rightarrow$ \textbf{sv} & $\rightarrow$ \textbf{syn}  & Avg \\
				\hline
				\hline
				
				\multirow{5}{*}{\begin{tabular}[c]{@{}c@{}}Single\\Best\end{tabular} } 
				& Source-only  & 59.2$\pm$0.6  & 97.2$\pm$0.6  & 84.7$\pm$0.8  &  77.7$\pm$0.8 & 85.2$\pm$0.6 & 80.8 \\ 
				&DAN~\cite{dan} & 63.8$\pm$0.7 & 96.3$\pm$0.5 & 94.2$\pm$0.9 & 62.5$\pm$0.7 & 85.4$\pm$0.8 & 80.4\\
				&CORAL~\cite{deepcoral} & 62.5$\pm$0.7 & 97.2$\pm$0.8 & 93.5$\pm$0.8 & 64.4$\pm$0.7 & 82.8$\pm$0.7 & 80.1 \\
				&DANN~\cite{dann} & 71.3$\pm$0.6 & 97.6$\pm$0.8 & 92.3$\pm$0.9 & 63.5$\pm$0.8 & 85.4$\pm$0.8 & 82.0\\
				&ADDA~\cite{adda} & 71.6$\pm$0.5 & 97.9$\pm$0.8 & 92.8$\pm$0.7 & 75.5$\pm$0.5 & 86.5$\pm$0.6 & 84.8\\
				
				\hline
				\multirow{6}{*}{ \begin{tabular}[c]{@{}c@{}}Source\\Combine\end{tabular} } 
				&Source-only & 63.4$\pm$0.7 & 90.5$\pm$0.8 & 88.7$\pm$0.9 & 63.5$\pm$0.9 & 82.4$\pm$0.6 &77.7\\	
				& DAN~\cite{dan}  &  67.9$\pm$0.8 & 97.5$\pm$0.6  & 93.5$\pm$0.8  &  67.8$\pm$0.6 & 86.9$\pm$0.5  & 82.7 \\
				& DANN~\cite{dann}  & 70.8$\pm$0.8  & 97.9$\pm$0.7  &  93.5$\pm$0.8 &  68.5$\pm$0.5 & 87.4$\pm$0.9  & 83.6 \\
				& JAN~\cite{JAN} & 65.9$\pm$0.7 & 97.2$\pm$0.7 & 95.4$\pm$0.8 & 75.3$\pm$0.7 & 86.6$\pm$0.6 &84.1  \\ 
				& ADDA~\cite{adda}  & 72.3$\pm$0.7  & 97.9$\pm$0.6  & 93.1$\pm$0.8  &  75.0$\pm$0.8 & 86.7$\pm$0.6  &  85.0 \\
				& MCD~\cite{max_discrepancy} & 72.5$\pm$0.7 & 96.2$\pm$0.8 & 95.3$\pm$0.7 & 78.9$\pm$0.8 & 87.5$\pm$0.7 & 86.1\\
				\hline
				
				\multirow{5}{*}{ \begin{tabular}[c]{@{}c@{}}Multi-\\Source\end{tabular} } 
				& MDAN~\cite{MDAN} &69.5$\pm$0.3& 98.0$\pm$0.9& 92.4$\pm$0.7& 69.2$\pm$0.6& 87.4$\pm$0.5 & 83.3 \\
				& DCTN~\cite{DCTN} & 70.5$\pm$1.2 & 96.2$\pm$0.8 & 92.8$\pm$0.3 & 77.6$\pm$0.4 & 86.8$\pm$0.8 & 84.8\\
				& $\rm M^{3}SDA$~\cite{M3SDA} & 72.8$\pm$1.1 & 98.4$\pm$0.7 & 96.1$\pm$0.8 & 81.3$\pm$0.9 & 89.6$\pm$0.6 &87.7\\	
				& MDDA~\cite{MDDA} &78.6$\pm$0.6& 98.8$\pm$0.4& 93.9$\pm$0.5& 79.3$\pm$0.8& 89.7$\pm$0.7 & 88.1 \\
				& LtC-MSDA & \textbf{85.6}$\pm$0.8 & \textbf{99.0}$\pm$0.4 & \textbf{98.3}$\pm$0.4 & \textbf{83.2}$\pm$0.6 & \textbf{93.0}$\pm$0.5 & \textbf{91.8} \\	
				
				\bottomrule[1.0pt]
			\end{tabular}
		\end{spacing}
	\end{table}
	
	
	\textbf{Performance comparison.} We compare our approach with state-of-the-art methods to verify its effectiveness. For the sake of fair comparison, we introduce three standards. (1) \emph{Single Best}: We report the best performance of single-source domain adaptation algorithm among all the sources. (2) \emph{Source Combine}: All the source domain data are combined into a single source, and domain adaptation is performed in a traditional single-source manner. (3) \emph{Multi-Source}: The knowledges learned from multiple source domains are transferred to target domain. For the first two settings, previous single-source UDA methods, \emph{e.g.} DAN~\cite{dan}, JAN~\cite{JAN}, DANN~\cite{dann}, ADDA~\cite{adda}, MCD~\cite{max_discrepancy}, are introduced for comparison. For the \emph{Multi-Source} setting, we compare our approach with four existing MSDA algorithms, MDAN~\cite{MDAN}, DCTN~\cite{DCTN}, $\rm M^{3}SDA$~\cite{M3SDA} and MDDA~\cite{MDDA}.
	
	
	\subsection{Experiments on Digits-five} \label{section4_2}
	
	\textbf{Dataset.} Digits-five dataset contains five digit image domains, including MNIST (\textbf{mt}) \cite{mnist}, MNIST-M (\textbf{mm}) \cite{dann}, SVHN (\textbf{sv}) \cite{svhn}, USPS (\textbf{up}) \cite{usps}, and Synthetic Digits (\textbf{syn}) \cite{dann}. 
	Each domain contains ten classes corresponding to digits ranging from 0 to 9. 
	We follow the setting in DCTN \cite{DCTN} to sample the data.
	
	\textbf{Results.} Table \ref{table_digit} reports the performance of our method compared with other works. Source-only denotes the model trained with only source domain data, which serves as the baseline. From the table, it can be observed that the proposed LtC-MSDA surpasses existing methods on all five tasks. In particular, a performance gain of 7.0\% is achieved on the ``$\rightarrow$ \textbf{mm}'' task. The results demonstrate the effectiveness of our approach on boosting model's performance through integrating multiple domains' knowledges. 
	
	
	\begin{table}[t]
		\begin{spacing}{1.03}
			\centering
			\scriptsize
			\caption{Classification accuracy (\%) on \emph{Office-31} dataset.} \label{table_office}
			\setlength{\tabcolsep}{4.0mm}
			\begin{tabular}{c|c|c|c|c|C{0.8cm}}
				\toprule[1.0pt]
				\multirow{1}{*}{Standards} & Methods & $\rightarrow$ D & $\rightarrow$ W & $\rightarrow$ A & Avg \\
				\hline
				\hline
				
				\multirow{5}{*}{\begin{tabular}[c]{@{}c@{}}Single\\Best\end{tabular} } 
				& Source-only  & 99.0  & 95.3  & 50.2  &  81.5 \\
				& RevGrad~\cite{revgrad}  & 99.2  & 96.4  &  53.4 & 83.0 \\
				& DAN~\cite{dan}  & 99.0  & 96.0  & 54.0  & 83.0 \\
				& RTN~\cite{RTN}  & \textbf{99.6}  & 96.8  & 51.0  & 82.5 \\
				& ADDA~\cite{adda}  & 99.4  & 95.3  & 54.6  & 83.1 \\
				\hline	
				
				\multirow{6}{*}{ \begin{tabular}[c]{@{}c@{}}Source\\Combine\end{tabular} } 
				&Source-only & 97.1  &  92.0 &  51.6 &  80.2\\
				&DAN~\cite{dan} & 98.8  &  96.2 &  54.9 &  83.3\\	
				&RTN~\cite{RTN} & 99.2  &  95.8 &  53.4 &  82.8\\
				&JAN~\cite{JAN} & 99.4  &  95.9 &  54.6 &  83.3 \\
				&ADDA~\cite{adda} & 99.2  &  96.0 &  55.9 & 83.7\\
				&MCD~\cite{max_discrepancy} & 99.5  &  96.2 &  54.4 &  83.4\\
				
				\hline	
				\multirow{5}{*}{ \begin{tabular}[c]{@{}c@{}}Multi-\\Source\end{tabular} }  
				& MDAN~\cite{MDAN}  & 99.2  & 95.4  &  55.2 &  83.3  \\	
				&DCTN~\cite{DCTN} & \textbf{99.6} & 96.9 & 54.9 & 83.8 \\	
				&$\rm M^{3}SDA$~\cite{M3SDA}& 99.4 & 96.2 & 55.4 & 83.7 \\			
				& MDDA~\cite{MDDA}   & 99.2  & 97.1 &  56.2 &  84.2  \\	
				& LtC-MSDA & \textbf{99.6} & \textbf{97.2} & \textbf{56.9} & \textbf{84.6} \\	
				
				\bottomrule[1.0pt]
			\end{tabular}
		\end{spacing}
	\end{table}
	
	
	\subsection{Experiments on Office-31} \label{section4_3}
	
	\textbf{Dataset.} Office-31 \cite{office_31_dataset} is a classical domain adaptation benchmark with 31 categories and 4652 images. It contains three domains: Amazon (A), Webcam (W) and DSLR (D), and the data are collected from office environment.
	
	\textbf{Results.} In Table \ref{table_office}, we report the performance of our approach and existing methods on three tasks. The LtC-MSDA model outperforms the state-of-the-art method, MDDA \cite{MDDA}, with 0.4\% in the term of average classification accuracy, and a 0.7\% performance improvement is obtained on the hard-to-transfer task, ``$\rightarrow$ A''. On this dataset, our approach doesn't have obvious superiority, which probably ascribes to two reasons. (1) First, domain adaptation models exhibit saturation when evaluated on ``$\rightarrow$ D'' and ``$\rightarrow$ W'' tasks, in which Source-only models achieve performance higher than 95\%. (2) Second, the Webcam and DSLR domains are highly similar, which restricts the benefit brought by multiple domains' interaction in our framework, especially in ``$\rightarrow$ A'' task.
	
	
	\begin{table*}[t]
		\begin{spacing}{1.01}
			\centering
			\scriptsize
			\caption{Classification accuracy (mean $\pm$ std \%) on \emph{DomainNet} dataset.} \label{table_domainnet}
			\setlength{\tabcolsep}{0.65mm}
			\begin{tabular}{c|c|c|c|c|c|c|c|C{0.8cm}}
				\toprule[1.0pt]
				\multirow{1}{*}{Standards} & Methods & $\rightarrow$ clp & $\rightarrow$ inf &$\rightarrow$ pnt & $\rightarrow$ qdr & $\rightarrow$ rel & $\rightarrow$ skt  & Avg \\
				\hline
				\hline
				
				\multirow{6}{*}{\begin{tabular}[c]{@{}c@{}}Single\\Best\end{tabular} } 
				& Source-only & 39.6$\pm$0.6 & 8.2$\pm$0.8 & 33.9$\pm$0.6 & 11.8$\pm$0.7 & 41.6$\pm$0.8 & 23.1$\pm$0.7 & 26.4\\
				&DAN~\cite{dan} &  39.1$\pm$0.5 & 11.4$\pm$0.8 & 33.3$\pm$0.6 & 16.2$\pm$0.4 & 42.1$\pm$0.7  & 29.7$\pm$0.9 & 28.6\\
				&JAN~\cite{JAN} &  35.3$\pm$0.7 & 9.1$\pm$0.6 & 32.5$\pm$0.7 & 14.3$\pm$0.6& 43.1$\pm$0.8  & 25.7$\pm$0.6 & 26.7\\
				&DANN~\cite{dann} &  37.9$\pm$0.7 & 11.4$\pm$0.9 & 33.9$\pm$0.6 & 13.7$\pm$0.6& 41.5$\pm$0.7  & 28.6$\pm$0.6 & 27.8\\
				&ADDA~\cite{adda} &  39.5$\pm$0.8 & 14.5$\pm$0.7 & 29.1$\pm$0.8 & 14.9$\pm$0.5& 41.9$\pm$0.8  & 30.7$\pm$0.7 & 28.4\\
				&MCD~\cite{max_discrepancy} & 42.6$\pm$0.3 &  19.6$\pm$0.8  & 42.6$\pm$1.0 & 3.8$\pm$0.6 &50.5$\pm$0.4&33.8$\pm$0.9& 32.2 \\
				\hline
				
				\multirow{6}{*}{ \begin{tabular}[c]{@{}c@{}}Source\\Combine\end{tabular} } 
				&Source-only & 47.6$\pm$0.5  & 13.0$\pm$0.4 & 38.1$\pm$0.5   & 13.3$\pm$0.4 & 51.9$\pm$0.9 & 33.7$\pm$0.5 & 32.9\\
				&DAN~\cite{dan}& 45.4$\pm$0.5&	12.8$\pm$0.9&	36.2$\pm$0.6&	15.3$\pm$0.4&	48.6$\pm$0.7&	34.0$\pm$0.5&	32.1  \\
				&JAN~\cite{JAN}& 40.9$\pm$0.4&	11.1$\pm$0.6& 35.4$\pm$0.5&	12.1$\pm$0.7&	45.8$\pm$0.6&	32.3$\pm$0.6&	29.6  \\
				&DANN~\cite{dann}& 45.5$\pm$0.6& 13.1$\pm$0.7&	37.0$\pm$0.7&	13.2$\pm$0.8&	48.9$\pm$0.7&	31.8$\pm$0.6& 32.6  \\
				&ADDA~\cite{adda}& 47.5$\pm$0.8& 11.4$\pm$0.7&	36.7$\pm$0.5&	14.7$\pm$0.5&	49.1$\pm$0.8&	33.5$\pm$0.5& 32.2  \\
				&MCD~\cite{max_discrepancy}& 54.3$\pm$0.6&	22.1$\pm$0.7&	45.7$\pm$0.6&	7.6$\pm$0.5&	58.4$\pm$0.7&	43.5$\pm$0.6& 38.5  \\
				
				\hline	
				\multirow{5}{*}{ \begin{tabular}[c]{@{}c@{}}Multi-\\Source\end{tabular} } 
				& MDAN~\cite{MDAN} &52.4$\pm$0.6& 21.3$\pm$0.8& 46.9$\pm$0.4& 8.6$\pm$0.6& 54.9$\pm$0.6& 46.5$\pm$0.7& 38.4 \\	
				&DCTN~\cite{DCTN} &48.6$\pm$0.7 & 23.5$\pm$0.6  &48.8$\pm$0.6  &7.2$\pm$0.5& 53.5$\pm$0.6 & 47.3$\pm$0.5 & 38.2 \\
				&$\rm M^{3}SDA$~\cite{M3SDA} &58.6$\pm$0.5& 26.0$\pm$0.9& 52.3$\pm$0.6& 6.3$\pm$0.6& 62.7$\pm$0.5& 49.5$\pm$0.8& 42.6 \\
				& MDDA~\cite{MDDA} &59.4$\pm$0.6& 23.8$\pm$0.8& 53.2$\pm$0.6& 12.5$\pm$0.6& 61.8$\pm$0.5& 48.6$\pm$0.8& 43.2 \\
				& LtC-MSDA & \textbf{63.1}$\pm$0.5 & \textbf{28.7}$\pm$0.7 & \textbf{56.1}$\pm$0.5 & \textbf{16.3}$\pm$0.5 & \textbf{66.1}$\pm$0.6 & \textbf{53.8}$\pm$0.6 & \textbf{47.4} \\	
				
				\bottomrule[1.0pt]
			\end{tabular}
		\end{spacing}
	\end{table*}
	
	
	\subsection{Experiments on DomainNet} \label{section4_4}
	
	\textbf{Dataset.} DomainNet \cite{M3SDA} is by far the largest and most difficult domain adaptation dataset. It consists of around 0.6 million images and 6 domains: clipart (clp), infograph (inf), painting (pnt), quickdraw (qdr), real (rel) and sketch (skt). Each domain contains the same 345 categories of common objects. 
	
	\textbf{Results.} The results of various methods on DomainNet are presented in Table \ref{table_domainnet}. Our model exceeds existing works with a notable margin on all six tasks. In particular, a 4.2\% performance gain is achieved on mean accuracy. The major challenges of this dataset are two-fold. (1) Large domain shift exists among different domains, \emph{e.g.} from real images to sketches. (2) Numerous categories increase the difficulty of learning discriminative features. Our approach tackles these two problems as follows. For the first issue, the global term of \emph{Relation Alignment Loss} constrains the similarity between two arbitrary categories to be consistent on all domains, which encourages better feature alignment in the latent space. For the second issue, the local term of \emph{Relation Alignment Loss} promotes the compactness of the same categories' features, which eases the burden of feature separation among different classes. 
	
	\section{Analysis} \label{sec5}
	
	In this section, we provide more in-depth analysis of our method to validate the effectiveness of major components, and both quantitative and qualitative experiments are conducted for verification. 
	
	\subsection{Ablation Study} \label{sec5_1}
	
	\textbf{Effect of domain adaptation losses.} In Table \ref{table_ablation_a}, we analyze the effect of global and local \emph{Relation Alignment Loss} on Digits-five dataset. On the basis of baseline setting (1st row), the global consistency loss (2nd rows) can greatly promote model's performance by promoting category-level domain alignment. For the local term, after adding it to the baseline configuration (3rd row), a 2.12\% performance gain is achieved, which demonstrates the effectiveness of $\mathcal{L}^{local}_{RAL}$ on enhancing the separability of feature representations. Furthermore, the combination of $\mathcal{L}^{global}_{RAL}$ and $\mathcal{L}^{local}_{RAL}$ (4th row) obtains the best performance, which shows the complementarity of global and local constraints. 
	
	\begin{table}[t]
		\begin{spacing}{1.0}
			\centering
			\caption{Ablation study for domain adaptation losses on global and local levels.} \label{table_ablation_a}
			\small
			\setlength{\tabcolsep}{2.3mm}
			\begin{tabular}{cc|ccccc|c}
				\toprule[1.0pt]
				{\tiny $\mathcal{L}^{global}_{RAL}$} & {\tiny $\mathcal{L}^{local}_{RAL}$} & $\rightarrow$ \textbf{mm} & $\rightarrow$ \textbf{mt} &$\rightarrow$ \textbf{up} & $\rightarrow$ \textbf{sv} & $\rightarrow$ \textbf{syn}  & Avg \\ 
				\hline
				&   & 74.85 & 98.60 & 97.95 & 74.56 & 88.54 & 86.90 \\
				$\checkmark$ &   & 82.49 & 98.97 & 98.06 & 81.64 & 91.70 & 90.57  \\
				&  $\checkmark$ & 79.57 & 98.64 & 98.06 & 78.66 & 90.16 & 89.02 \\
				$\checkmark$ &  $\checkmark$ & 85.56 & 98.98 & 98.32 & 83.24 & 93.04 & 91.83 \\
				\bottomrule[1.0pt]		
			\end{tabular}
		\end{spacing}
		\vspace{-3mm}
	\end{table}
	
	\begin{table}[t]
		\begin{spacing}{1.0}
			\centering
			\caption{Ablation study for three kinds of classification losses.} \label{table_ablation_b}
			\small
			\setlength{\tabcolsep}{1.6mm}
			\begin{tabular}{ccc|ccccc|c}
				\toprule[1.0pt]
				$\mathcal{L}^{src}_{cls}$ \, & $\mathcal{L}^{proto}_{cls}$ & $\mathcal{L}^{tgt}_{cls}$ & $\rightarrow$ \textbf{mm} & $\rightarrow$ \textbf{mt} &$\rightarrow$ \textbf{up} & $\rightarrow$ \textbf{sv} & $\rightarrow$ \textbf{syn}  & Avg \\ 
				\hline 
				
				$\checkmark$&  &   & 73.65 & 98.47 & 96.61 & 78.20 & 88.93 & 87.17 \\ 
				$\checkmark$ & $\checkmark$  &  & 78.44 & 98.64 & 96.77 & 79.24 & 89.05 & 88.43 \\
				$\checkmark$&  & $\checkmark$  & 81.36 & 98.76 & 97.93 & 81.26 & 91.70 & 90.20 \\
				$\checkmark$ & $\checkmark$ & $\checkmark$ & 85.56 & 98.98 & 98.32 & 83.24 & 93.04 & 91.83 \\
				\bottomrule[1.0pt]
			\end{tabular}
		\end{spacing}
		\vspace{-3mm}
	\end{table}
	
	
	\textbf{Effect of classification losses.} Table \ref{table_ablation_b} presents the effect of different classification losses on Digits-five dataset.
	The configuration of using only source samples' classification loss $\mathcal{L}^{src}_{cls}$ (1st row) serves as the baseline. After adding the entropy constraint for target samples (3rd row), the accuracy increases by 3.03\%, which illustrates the effectiveness of $\mathcal{L}^{tgt}_{cls}$ on making target samples' features more discriminative. 
	Prototypes' classification loss $\mathcal{L}^{proto}_{cls}$ is able to further boost the performance by constraining prototypes' distinguishability (4th row).
	
	\subsection{Sensitivity Analysis} \label{sec5_2}
	
\begin{figure}[t]
		\centering
		\includegraphics[width=0.98\textwidth]{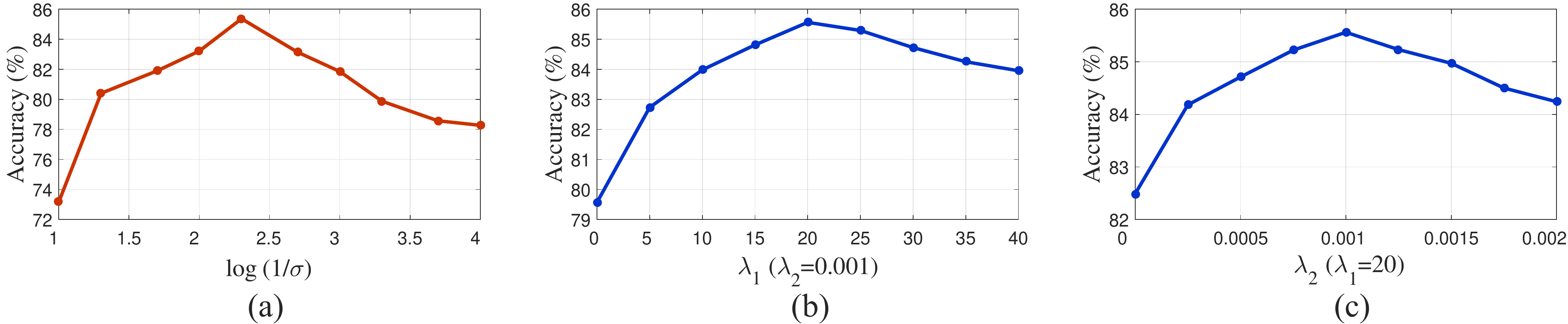}
		\caption{Sensitivity analysis of standard deviation $\sigma$ (left) and trade-off parameters $\lambda_1$, $\lambda_2$ (middle, right). (All results are reported on the ``$\rightarrow$ \textbf{mm}'' task.)} 
		\label{fig_sensitivity}
		\vspace{-2mm}
	\end{figure}

	\textbf{Sensitivity of standard deviation $\sigma$.} In this part, we discuss the selection of parameter $\sigma$ which controls the sparsity of adjacency matrix. In Figure \ref{fig_sensitivity}(a), we plot the performance of models trained with different $\sigma$ values. The highest accuracy on target domain is achieved when the value of $\sigma$ is around 0.005. Also, it is worth noticing that obvious performance decay occurs when the adjacency matrix is too dense or sparse, \emph{i.e.} $\sigma > 0.05$ or $\sigma < 0.0005$. 
	
	
	\begin{figure}[t]
		\subfloat[Adjacency matrix $\textbf{A}$ with and without RAL constraint.]{\label{fig:1}\includegraphics[width=.475\linewidth]{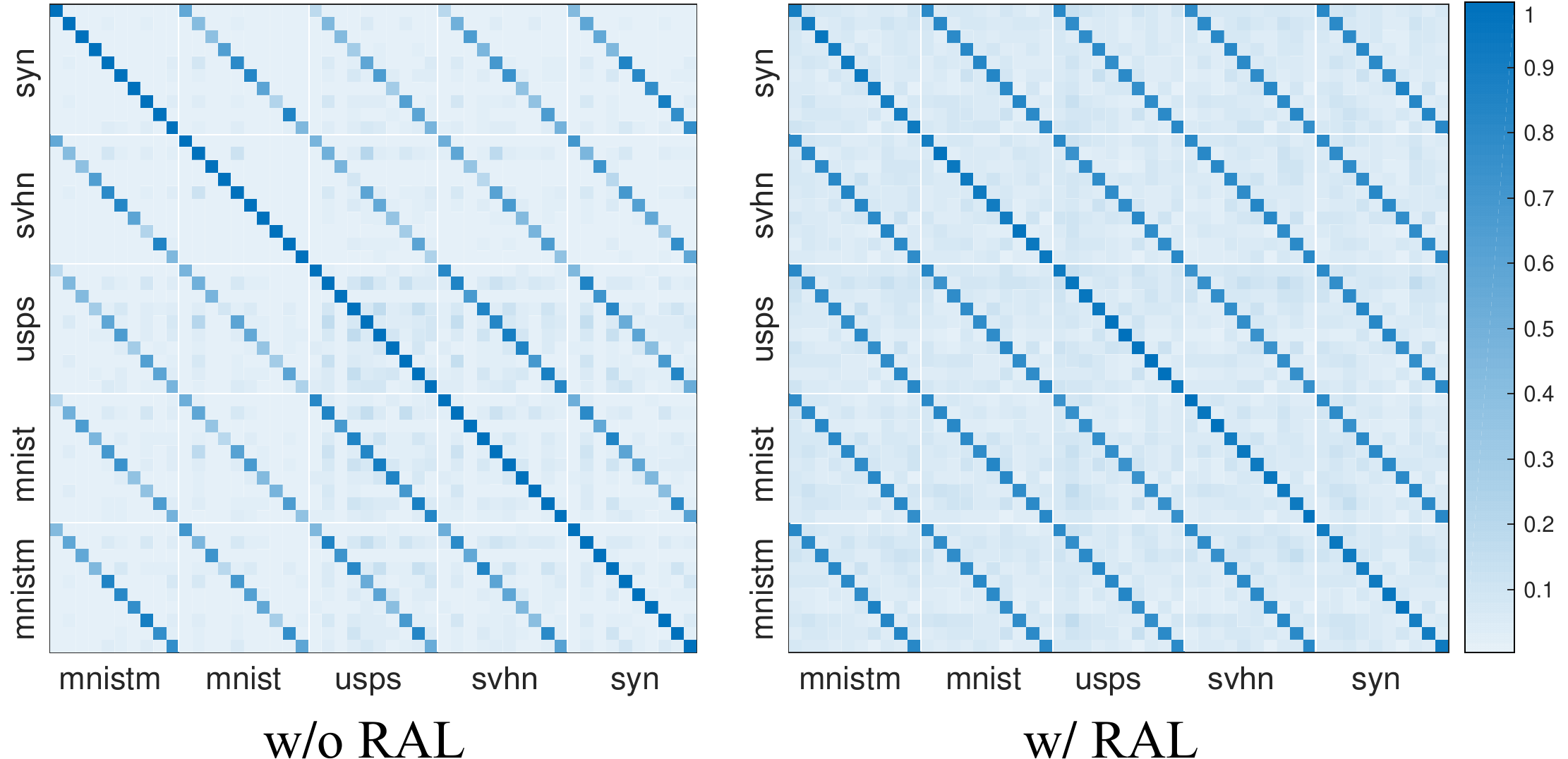}}
		\hfill
		\subfloat[Feature distributions of source domain (``blue'') and target domain (``red'').]{\label{fig:2}\includegraphics[width=0.50\linewidth]{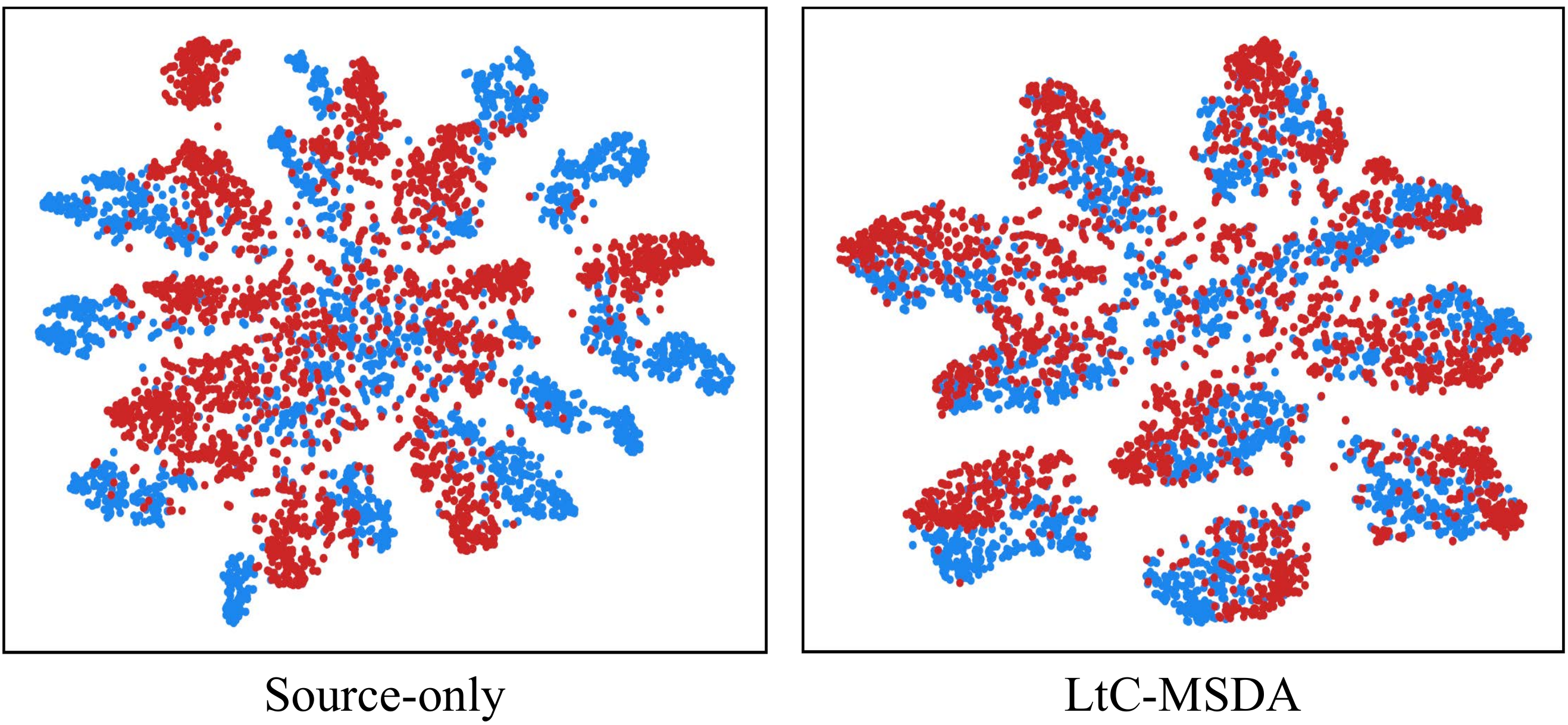}}
		\caption{Visualization of adjacency matrix and feature embeddings. 
			(All results are evaluated on the ``$\rightarrow$ \textbf{mm}'' task.)
		}
		\label{fig_visualization}
		\vspace{-2mm}
	\end{figure}
	
	
	\textbf{Sensitivity of trade-off parameters $\lambda_1$, $\lambda_2$.} In this experiment, we evaluate our approach's sensitivity to $\lambda_1$ and $\lambda_2$ which trade off between domain adaptation and classification losses. 
	Figure \ref{fig_sensitivity}(b) and Figure \ref{fig_sensitivity}(c) show model's performance under different $\lambda_1$ ($\lambda_2$) values when the other parameter $\lambda_2$ ($\lambda_1$) is fixed. From the line charts, we can observe that model's performance is not sensitive to $\lambda_1$ and $\lambda_2$ when they are around 20 and 0.001, respectively. In addition, performance decay occurs when these two parameters approach 0, which demonstrates that both global and local constraints are indispensable. 
	
	
	\subsection{Visualization} \label{sec5_3}
	
	\textbf{Visualization of adjacency matrix.} Figure \ref{fig_visualization}(a) shows the adjacency matrix $\textbf{A}$ before and after applying the \emph{Relation Alignment Loss} (RAL), in which each pixel denotes the relevance between two categories from arbitrary domains. It can be observed that, after adding RAL, the relevance among various categories is apparently more consistent across different domains, which is compatible with the relational structure constrained by the global term of RAL.
	
	
	\textbf{Visualization of feature embeddings.} In Figure \ref{fig_visualization}(b), we utilize t-SNE \cite{tsne} to visualize the feature distributions of one of source domains (SVHN) and target domain (MNIST-M). 
	Compared with the Source-only baseline, the proposed LtC-MSDA model makes the features of target domain more discriminative and better aligned with those of source domain. 
	
	
	\section{Conclusion} \label{sec6}
	
	In this paper, we propose a Learning to Combine for Multi-Source Domain Adaptation (LtC-MSDA) framework. In this framework, the knowledges learned from multiple domains are aggregated to assist the prediction for query samples. Furthermore, we conduct class-relation-aware domain alignment via constraining global category relationships and local feature compactness. 
	Extensive experiments and analytical studies demonstrate the prominent performance of our approach under various domain shift settings.
	
	
	\section*{Acknowledgement} \label{sec7}
	
	This work was supported by National Science Foundation of China (61976137, U1611461, U19B2035) and STCSM(18DZ1112300). Authors would like to appreciate the Student Innovation Center of SJTU for providing GPUs.
	
	\clearpage

	%
	%
	
	\bibliographystyle{splncs04}
	\bibliography{main.bib}

	\clearpage
	\newpage
	\section*{{\Large Appendices}}
	\label{sec_appendix}
	\appendix
	
\section{Detailed Experimental Setups}
In this part, we provide detailed experimental setups. 
For the sake of fair comparison, we follow the backbone setting in \cite{DCTN,M3SDA} for different tasks. 
In our framework, a feature vector is employed to update global prototypes and also serves as query samples, whose dimension varies as the backbone architecture.
For different tasks, we list the basic training settings in Table \ref{exp_setup}. 

\vspace{-0.6cm}
\begin{table}[!htb]
	\begin{spacing}{1.2}
		\centering
		\scriptsize
		\caption{The experimental setups in three different tasks.} \label{exp_setup}
		\setlength{\tabcolsep}{1.2mm}
		\begin{threeparttable}
			\begin{tabular}{c|c|c|c|c|c|c|c|c}
				\toprule[1.0pt]
				
				\multirow{2}{*}{dataset} & \multirow{2}{*}{domains} & \multirow{2}{*}{classes} & \multirow{2}{*}{\begin{tabular}[c]{@{}c@{}}image\\ size\end{tabular}} & \multirow{2}{*}{backbone} &
				\multirow{2}{*}{\begin{tabular}[c]{@{}c@{}}batch\\ size\tnote{*} \end{tabular}} & \multirow{2}{*}{\begin{tabular}[c]{@{}c@{}}learning\\ rate\end{tabular}} & \multirow{2}{*}{\begin{tabular}[c]{@{}c@{}}training\\  epoch\end{tabular}} & \multirow{2}{*}{\begin{tabular}[c]{@{}c@{}}feature \\ dimension\end{tabular}} \\
				&         &               &           &         & &   &  \\ 		
				\hline
				
				Digits-five & 5       & 10      & $32 \times 32$      & 3 conv-2 fc      & 128 & $2 \times 10^{-4}$  & 100            & 2048               \\
				Office-31\cite{office_31_dataset}  & 3       & 31      & $252 \times 252$    & AlexNet  & 16   & $5 \times 10^{-5}$ & 100            & 4096               \\
				DomainNet\cite{M3SDA}  & 6       & 345     & $224 \times 224$    & ResNet-101 & 16  & $5 \times 10^{-5}$ & 20             & 2048              \\
				PACS\cite{pacs}  & 4       & 7     & $224 \times 224$    & ResNet-18 & 16  & $5 \times 10^{-5}$ & 100             & 512              \\

				\bottomrule[1.0pt]
			\end{tabular}
			\begin{tablenotes}
				\scriptsize
				\item[*] Batch size here denotes the number of examples sampled from one domain in each iteration.
			\end{tablenotes}
		\end{threeparttable}
	\end{spacing}
	\vspace{-10mm}
\end{table}


\begin{figure}[ht]
	\centering
	\setlength{\belowcaptionskip}{-0.4cm}
	\includegraphics[width=0.65\textwidth]{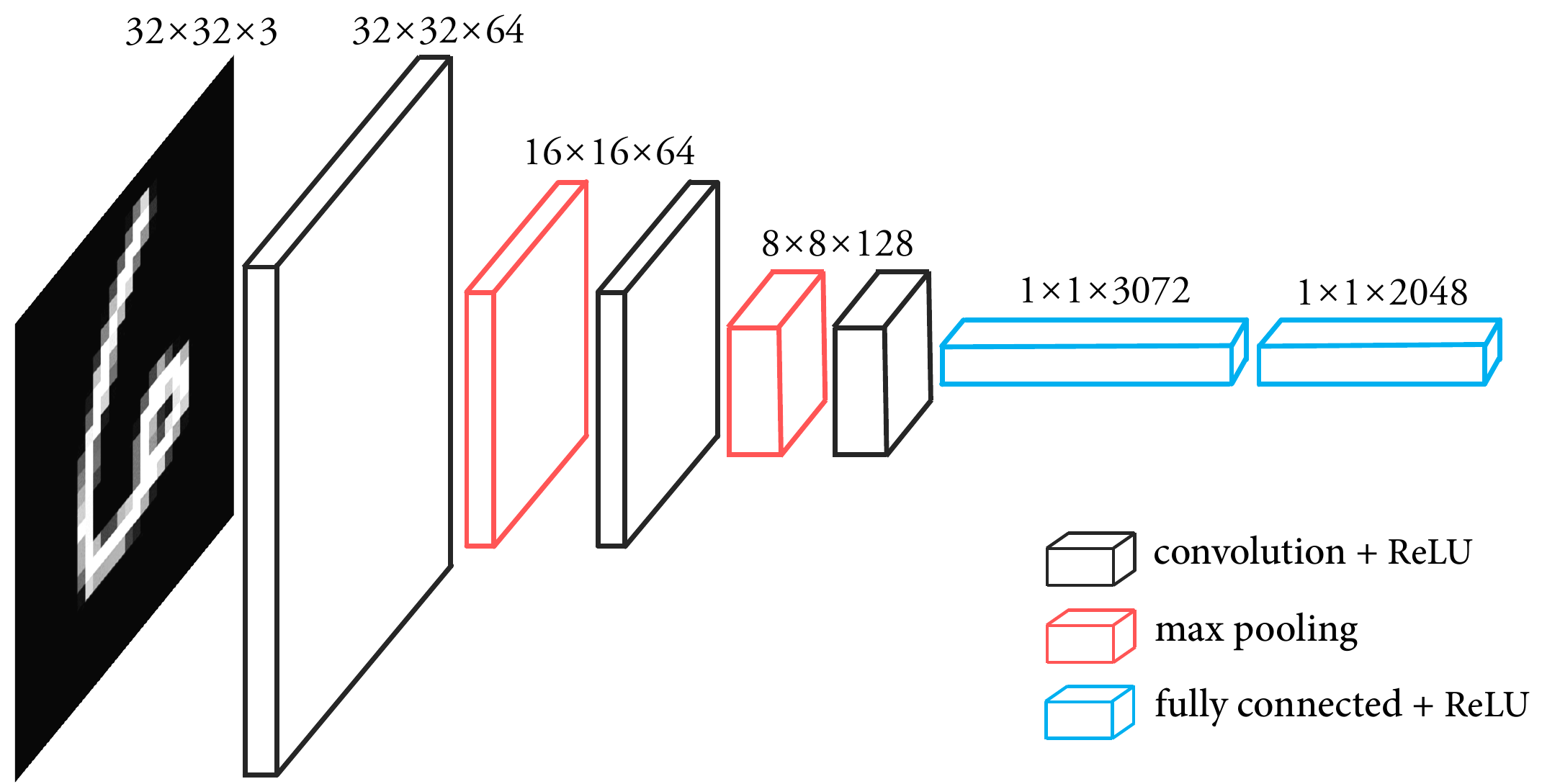}
	\caption{The network architecture for experiments on Digits-five dataset.} 
	\label{fig_architecture}
\end{figure}

In Figure \ref{fig_architecture}, we provide the detailed network architecture for digits experiments, which mainly follows the design in \cite{M3SDA}. Inputted with an image whose spatial size is $32 \times 32$, three convolution-based modules produce an $8 \times 8 \times128$ feature map. After that, the feature map is flattened, and a 2048-dimensional feature vector is generated by two fully connected layers. 

\section{Experiments on PACS} \label{supp_sec2}

\textbf{Dataset.} PACS \cite{pacs} dataset contains 4 domains, \emph{i.e.} Photo (P), Art paintings (A), Cartoon (C) and Sketch (S). Each domain contains 7 categories, and significant domain shift exists between different domains.

\textbf{Results.} Table \ref{table_pacs} reports the performance of our method compared with other works. Source-only denotes the model trained with only source domain data, which serves as the baseline. As shown in the table, the proposed LtC-MSDA model achieves the highest accuracy on all four tasks of PACS dataset, and a 3.74\% performance gain is obtained in the term of average accuracy.
By combining the knowledges learned from multiple domains, our model show superior performance under huge domain shift settings.

\begin{table}[t]
	\begin{spacing}{1.1}
		\centering
		\small
		\caption{Classification accuracy (\%) on \emph{PACS} dataset.} \label{table_pacs}
		\setlength{\tabcolsep}{3.6mm}
		\begin{tabular}{c|cccc|c}
			\toprule[1.0pt]
			Methods & $\rightarrow$ A & $\rightarrow$ C & $\rightarrow$ S & $\rightarrow$ P & Avg \\
			\hline
			Source-only  & 75.97 & 73.34 & 64.23 & 91.65 & 76.30 \\
			MDAN~\cite{MDAN}  & 83.54 & 82.34 & 72.42 & 92.91 & 82.80 \\
			DCTN~\cite{DCTN}  & 84.67 & 86.72 & 71.84 & 95.60 & 84.71 \\
			$\rm M^{3}SDA$~\cite{M3SDA}  & 84.20 & 85.68 & 74.62 & 94.47 & 84.74 \\
			MDDA~\cite{MDDA}  & 86.73 & 86.24 & 77.56 & 93.89 & 86.11 \\
			LtC-MSDA  & \textbf{90.19} & \textbf{90.47} & \textbf{81.53} & \textbf{97.23} & \textbf{89.85} \\		
			\bottomrule[1.0pt]
		\end{tabular}
	\end{spacing}
\end{table}

\end{document}